\pdfoutput=1

\documentclass[11pt]{article}

\usepackage[]{acl}

\usepackage{times}
\usepackage{latexsym}

\usepackage[T1]{fontenc}

\usepackage[utf8]{inputenc}

\usepackage{microtype}
\usepackage[ruled]{algorithm2e} 

\SetAlFnt{\small}
\SetAlCapFnt{\small}
\SetAlCapNameFnt{\small}
\SetAlCapHSkip{0pt}
\IncMargin{-\parindent}

\usepackage[normalem]{ulem}
\usepackage{amsmath}
\usepackage{graphicx}
\usepackage{mathtools}
\usepackage{caption}
\usepackage{subcaption}
\usepackage{mwe}
\usepackage{bm}
\usepackage{booktabs}
\usepackage{multirow}
\usepackage{nicefrac}
\usepackage{makecell}

\title{Probing Causes of Hallucinations in Neural Machine Translations}

\author{%
  Jianhao Yan \\
  WeChat AI, Tencent, China \\
  \texttt{elliottyan37@gmail.com} \\
  \And
  Fandong Meng \\
  WeChat AI, Tencent, China \\
  \AND
  Jie Zhou \\
  WeChat AI, Tencent, China \\
  
}
\begin{document}
\maketitle
\begin{abstract}
Hallucination, one kind of the pathological translations that bothers Neural Machine Translation, has recently drawn much attention. In simple terms, hallucinated translations are fluent sentences but barely related to source inputs. 
Arguably, it remains an open problem how hallucination occurs. 
In this paper, we propose to use probing methods to investigate the causes of hallucinations from the perspective of model architecture, aiming to avoid such problem in future architecture designs. By conducting experiments over various NMT datasets, we find that hallucination is often accompanied by the deficient encoder, especially embeddings, and vulnerable cross-attentions, while, interestingly, cross-attention mitigates some errors caused by the encoder.
\end{abstract}

\section{Introduction}
Huge success has been achieved in recent years for Neural Machine Translation (NMT) by applying techniques like recursive neural networks, and self-attention networks \cite{bahdanau2014neural,vaswani2017attention}. However, current NMT architectures are prone to inadequate translations and wrong named-entity translations and etc.. Hallucination \cite{leehallucinations,wang2020exposure,raunak2021curious} lies on the extreme end among these problems, which is identified by pathological translations that are barely related to source sentences. 

In order to understand the causes of hallucination, previous work mainly analyzes the question from the perspective of mismatches between training and testing.
Many insightful phenomenons are observed. 
For instance, perturbed input sentences, domain shift, and exposure bias are shown to be highly correlated with the generation of such problematic translations.  

Nevertheless, previous work generally examines hallucination by viewing models as a black-box, which completely ignores the roles different network components play in hallucination.
In this work, we fill in the blank by employing probing \cite{adi2016fine,hupkes2018visualisation} techniques to understand how each network component performs when the hallucination happens.
In particular, we probe the state-of-the-art Transformer models following \citet{xu2021probing}, and propose to measure the difference in word translation accuracies between hallucinated and un-hallucinated scenarios. 
We select inputs from in-domain and out-of-domain test sets, respectively, as domain shift is one of the main reasons for generating hallucination. By investigating different components of the model architecture, we can find which part of the model suffers the most in this problem. 

Extensive experiments are conducted over various NMT benchmarks across different domains and languages. The deficient encoder layers, especially the embedding layer, and cross-attentions have been proven to be highly correlated with the occurrence of hallucination. The decoder self-attentions are not susceptible to hallucination. Interestingly, the cross-attentions provide good inductive biases that help mitigate errors caused by encoder layers. 
\section{Background}
\subsection{Neural Machine Translation}
In this section, we briefly describe how an NMT model works. 
Given a source sentence $x$ and a target sentence $y$, an NMT model $M$ predicts the probability for a certain target token $y_t$ conditioned on previous generated tokens $y_{1:t-1}$,
\begin{gather}
    P(y_t|x, y_{1:t-1}) = M(x, y_{1:t-1}),
\end{gather}
where $t$ represents the target side time step. 

The training objective of an NMT model is usually the cross-entropy loss defined over reference $\hat{y}$,
\begin{gather}
    \mathcal{L} = - \sum_{t} \log P(\hat{y}_t|x, \hat{y}_{1:t-1}).
\end{gather}

\subsection{Hallucination}
Hallucination is characterized by the disentangled translation with the source input. 
\citet{raunak2021curious} further categorize hallucination into two kinds, \emph{Hallucination under Perturbations} (HP) and \emph{Natural Hallucination} (NH). Under perturbation, the hallucinated output is identified by the adjusted BLEU \cite{leehallucinations} between generated translations for perturbed and un-perturbed inputs, where the adjusted BLEU is a re-weighted BLEU score only considering 1-grams and 2-grams. 
As for NH, human detection is commonly used to detect hallucinated samples, and it is an impediment for fast experiment cycles \cite{raunak2021curious}. 
In this paper, we extend the method used in HP and apply adjusted BLEU to identify NP, except for comparing reference with generated outputs. Our hallucination detection algorithm is Algorithm \ref{alg:hallu_det}. The 0.01 adjusted-bleu threshold comes from \citet{leehallucinations}, which performs quite well in our preliminary experiments.
We use this algorithm to select hallucinated samples in the following contents. 

\begin{algorithm}[t]

\SetAlgoLined
\SetKwInOut{Input}{Input}
\SetKwInOut{Output}{Output}
\Input{NMT Model, Parallel Corpus $(X, Y)$}
\Output{Hallucinated Samples $H$}

\SetKwFunction{FDFS}{dfsTopK}

\For{$x, y \in (X, Y)$}{
    
    $y' = Beam\_Search(Model, x)$
    
    \If{$adjusted\_bleu(y', y) < 0.01$}{
        add $x$ to $H$
    }
}
\caption{Natural Hallucination Detection.}
\label{alg:hallu_det}
\end{algorithm}

\section{Probing Methods}
This section describes how we probe each layer in Transformer models using the word translation task. 
Given a converged NMT model $M$, we freeze and extract each layer's representation to predict the next word. The performance of the word translation task is a good proxy for evaluating the translation capability of the model's internal representations. 

\subsection{Encoder Layers}
For each encoder layer in a Transformer model, the hidden states are passed through a self-attention layer and an FFN layer \cite{vaswani2017attention},
\begin{align}
    s_i &= \text{Self\_Att}_{i}(h_{i-1}) + h_{i-1}, \\
    h_i &= \text{FFN}_i(h_{i-1}) + s_{i}.
\end{align}

Then, to analyze the ability of word translations for encoder layers, it is necessary to align the encoder representations to target side orders. Following \citet{xu2021probing}, we extract all cross-attention matrices and compute the weighted sum over them to gather a transformation matrix $\hat{A}$,
\begin{align}
    \hat{A} = \sum_{j=1}^{d*k} A_j * p_j, ~p_j = \frac{e^{w_j}}{\sum^{d*k}_{k=1}e^{w_k}},
\end{align}
where $d$ and $k$ denotes the number of decoder layers and attention heads, and $w$ is a trainable weight vector. 
Next, we align each layer's representation $s_i/h_i$ into target side with $T$, and pass through a softmax layer for predictions:
\begin{align}
    t_i = \hat{A}^T \times h_i,
    P_i(y) = \text{Softmax}(W^T t_i).
\end{align}

This method is minimally invasive in that only linear projection $W$ and weight vector $w$ are trainable while the original parameters of the Transformer model are frozen.  

\subsection{Decoder Layers}
As for decoder layers, it is unnecessary to align the tokens to the target side, and we can use the shifted target sequence to compute word translation accuracy. Specifically, we take out the probed layer representations and multiply them with the classification heads of the original Transformer model. As discussed in \cite{li-etal-2019-word,xu2021probing}, self-attention and cross-attention modules in Transformer decoder play different roles in translation process. Therefore, aside from the standard probing results with complete testset and hallucinated samples, we also conduct word translation experiments without the current layer's self-attention module or cross-attention module. 
\section{Experiments and Findings}
In our experiments, we only study NH, as HP deliberately attacks the NMT model and plunges results into extreme cases, hindering understanding of hallucinations' causes. We use domain shift to generate NH samples in our experiments, where we use WMT datasets (News) as our in-domain sets and other domains as out-of-domain sets.
\subsection{Experimental Details}
Here, we first describe our experimental details. The training set we use is the standard WMT'14 English-German dataset, containing 4.5M preprocessed sentence pairs. We apply tokenization and use byte pair encoded (BPE) with 32K merge operations to split each sentence pair. The vocabularies of English and German are shared, and we choose \emph{newstest2013} as the validation set and \emph{newstest2014} as the test set, which contain 3000 and 3003 sentences, respectively.

For our out-of-domain datasets, we use the IWSTL 2014 testset \cite{cettolo2014report} which contains spoken parallel sentence pairs from TED talks.
The valid set is randomly selected from the training set, which contains 8417 sentences. The test set combines IWSLT test sets from 2010 to 2014, containing 7883 sentences. The statistics of identified hallucination of datasets can be found in Sec. \ref{sec:stat}.

We use the Transformer-base through all our experiments.
All hyper-parameter settings are the same with \citet{vaswani2017attention}. Our baseline model for WMT'14 English-German scores 27.3 in terms of BLEU.
For newly added probing parameters, we use the $512 \times 512$ projection matrix. 
All our probing models are trained with 100k steps with 25k tokens per batch, and we average the last five checkpoints for testing. 

\begin{table}
\centering
\setlength\tabcolsep{2.5pt}
\begin{tabular}{c||c |c |c |c |c |c }
\toprule
\multirow{2}{*}{\bf{Layer}} & \multicolumn{2}{c|}{\bf WMT'14}  & \multicolumn{2}{c|}{\bf IWSLT All}  & \multicolumn{2}{c}{\bf IWSLT Hallu.}\\
\cmidrule{2-7}
& {BLEU} & {Acc.} & {BLEU} & {Acc.} & {BLEU} & {Acc.} \\
\midrule
Emb. & 6.58 & 37.0 & 7.49	& 39.0 & 1.66 & 28.6 \\
\midrule
1 &7.14	&35.6 & 7.67 & 38.0 & 1.94 & 28.4 \\
2 &7.47	&37.0 & 7.88 & 39.3 & 2.04 & 29.6 \\
3 &7.66 & 38.2 & 8.33 & 40.3 & 1.98 & 30.1 \\
4 & 8.55&39.7 &9.07 & 41.8 & 2.18 & 31.0 \\
5 & 9.15&40.9 &9.80 & 43.0 & 2.46 & 31.2 \\
6 & 9.44 & 41.4 & 10.22 & 43.4 & 2.53 & 31.5 \\
\bottomrule
\end{tabular}
\caption{"\textbf{Hallu.}" stands for the hallucination parts of IWSLT, and "\textbf{All}" denotes the complete IWSLT testsets. "Acc." denotes accuracy and ranges from 0\% to 100\%.}
\vspace{-13pt}
\label{table:iwslt}
\end{table}

\begin{table*}[th]
\centering
\setlength\tabcolsep{5pt}
\begin{tabular}{c||c | c| c | r | c | c | r | c | c |r  }
\toprule
\multirow{2}{*}{\bf{Layer}} & \multicolumn{4}{c|}{\bf Standard}  & \multicolumn{3}{c|}{\bf No Self-Att}  & \multicolumn{3}{c}{\bf No Cross-Att} \\
\cmidrule{2-11}
& WMT & {All} & {Hallu.} & $\Delta$ & {All} & {Hallu.}  & $\Delta$ & {All} & {Hallu.}  & $\Delta$ \\
\midrule
1 & 0.0 & 0.0 & 0.0 & +0.0 & 0.5 & 0.8 & +0.3& 0.1 & 0.0 &+0.0\\
2 & 3.4 & 4.1 & 6.6  & +2.5 & 7.5 & 9.1& +1.6& 8.2 & 10.4& +2.2\\
3 & 13.1 & 13.6 & 12.8 & -0.8& 15.5 & 13.0 &-2.5& 13.4 & 14.1& +0.7 \\
4 & 32.6 & 35.3 & 29.5 & -5.8& 29.1 & 24.3 &-4.8& 20.3 & 20.4 & +0.1\\
5 & 58.3 & 59.0 & 47.7 & -11.3& 35.8 & 24.2 & -11.6& 23.3 & 21.5 &-1.8\\
6 & 69.4 & 68.0 & 55.3 & -12.7 & 23.4 & 17.0 & -6.4& 5.4 & 4.6 &-0.8\\
\bottomrule
\end{tabular}
\caption{Decoder layer experiments for word translation over hallucinations. We test over three settings. "Standard" means the standard transformer. "No Self-Att" means removing the current layer's self-attention module. "No Cross-Att" means removing the current layer's cross-attention module. $\Delta$ denotes the corresponding difference between each "All" and "Hallu.".}
\label{table:dec}
\vspace{-11pt}
\end{table*}

\subsection{IWSLT Results}
\label{sec:iwslt}
We list the word translation results of probing in Table \ref{table:iwslt}. We report BLEU scores and accuracies on both the complete IWSLT test set and its hallucinated parts to understand the critical component of hallucination. We also list the results on WMT'14 English-German testsets, in order for readers to better interpret these numbers.

When investigating these results, we draw several interesting conclusions: 
Firstly, our probing method with encoder hidden can achieve about 41.4\% accuracy, showing strong word translation capability. The embedding itself can achieve 37\% word translation accuracy, which indicates the model learns to translate starting from the level of word embeddings. 
These findings are consistent with results in \citet{xu2021probing}.

Secondly, we find "IWSLT All" has good accuracy, about 39\% for encoder embedding and 43.4\% for encoder outputs, where encoder representation in "WMT'14" only predicts with 41.4\% accuracy. 
Even though the results of "WMT'14" and "IWSLT All" are not directly comparable, these numbers show that domain shift does not necessarily lead to bad translation capabilities. 

Thirdly, comparing "IWSLT All" with "IWSLT Hallu.", we can see that the absolute numbers of accuracy drop drastically, with more than 10\% decreases. Considering that they are of the relationship for the part and whole and are in the same domain, such decreases are quite surprising. 
The above findings bridge the performance decrease of word translation with hallucination and demonstrate that the encoder's capability becomes deficient in hallucination. 

Fourthly, by examining the results across each layer, we find the translation capabilities of encoder layers are not highly affected by hallucinated inputs.
Comparing "Emb." with "Layer 6", the accuracy for "IWSLT All" increases from 39.0\% to 43.4\% (+4.4), and the accuracy for "IWSLT Hallu" increases from 28.6\% to 31.5\% (+2.9). 
The improvements only show a minor decrease, which is counter-intuitive. We will revisit this in Sec. \ref{sec:w/o_cross}.

Then what is the cause of hallucination? We find the most of the decreases comes from the embedding level. 
Without any participation of encoder layers, the embedding layer accounts for a decrease of about 10\% accuracy, which proves \textbf{hallucination is closely connected to the embedding layer.}

\subsection{Analysis of Decoder Layers}
In this section, we provide our probing results for decoder layers. As shown in Table \ref{table:dec}, we conduct experiments over three settings, "standard Transformer" (Standard), "no self-attention at current layer"(No Self-Att), and "no cross-attention at current layer" (No Cross-Att). By comparing $\Delta$s between "Standard" and "No Self-Att", layers without self-attention show no sign of shrinking the gap between "All" and "Hallu.". It indicates that self-attention is not susceptible in hallucination. On the other hand, by comparing $\Delta$s between "Standard" and "No Cross-Att", the gaps between "All" and "Hallu" almost disappear, showing cross-attention is the most affected module in hallucination.  

\subsection{Without Cross Attentions}
\label{sec:w/o_cross}
In our encoder experiments, it is necessary to use encoder-decoder cross attention to perform alignment of encoder outputs. However, this brings in concerns about how these attentions affect our analysis, since we prove that cross-attentions are the mostly affected modules in decoder layers. In this section, we still conduct word translation experiments except that we do not use cross-attention anymore. Instead, we directly transform the analyzed layer outputs with our projection matrix and predict translation words. To this end, accuracy is not a suitable metric for us without alignment. Therefore, we use BLEU scores. As the scores are too small to show any difference, we mainly analyze our results with 1-gram BLEU scores. 

As shown in Table \ref{table:w/o_cross}, the embedding layer still is responsible for a huge decreases in hallucination, a decline of 7.3 1-BLEU. However, the gap between "All" and "Hallu." expands when increasing the depth of encoder layers. It demonstrates that hallucinated samples harm the translation capabilities of encoder layers. Recall that in Sec. \ref{sec:iwslt}, increasing depth does not affect much. Since the only difference for these two experiments is the involvement of cross-attentions, we posit that the cross-attention module has a strong inductive bias and mitigates the loss of translation capabilities in encoder layers. 

\begin{table}
\centering
\setlength\tabcolsep{3pt}
\begin{tabular}{c||c |c |c |c |r }
\toprule
\multirow{2}{*}{\bf{Layer}} &  \multicolumn{2}{c|}{\bf IWSLT All}  & \multicolumn{2}{c|}{\bf IWSLT Hallu.} & \multirow{2}{*}{\bf{$\Delta$}}\\
\cmidrule{2-5}
&  {BLEU} & {1-BLEU} & {BLEU}  & {1-BLEU}\\
\midrule
Emb.& 1.1 & 15.9 & 0.9 & 8.6 & -7.3 \\
1 & 3.8 & 32.5 & 1.5 & 16.1 & -16.3 \\
2 & 4.3 & 34.7 & 1.3 & 20.5 & -14.2 \\
3 & 5.4 & 37.5 & 1.6 & 22.0 & -15.5 \\
4 & 5.9 & 38.8 & 1.7 & 22.8 & -16.0 \\
5 & 6.7 & 40.5 & 1.9 & 24.0 & -16.5 \\
6 & 6.8 & 40.5 & 1.8 & 23.1 & -17.5 \\
\bottomrule
\end{tabular}
\caption{Word translation results without usage of cross attention layers. $\Delta$ denotes differences between 1-BLEU scores of "All" and "Hallu.".}
\label{table:w/o_cross}
\vspace{-10pt}
\end{table}

\subsection{Other Experiments}
To validate our findings also work in other domains and languages, we conduct experiments on two more settings. First, we use a diverse multi-domain MT dataset first proposed by \citet{koehn2017six} and re-processed by \citet{aharoni2020unsupervised}. Second, we additionally conduct English-French experiments on in-domain WMT'14 En-Fr and out-of-domain IWSLT'14 En-Fr datasets.
The results support our findings generalize across domains and languages.
Due to the space limitation, we put these results in the Appendix, and we refer our readers to Sec. \ref{appendix:md5} and \ref{appendix:en-fr}.

\section{Conclusion}
In this paper, we propose to use probing method to understand the cause of hallucinations, from the perspective of the model architecture. We analyze each layer of the state-of-the-art Transformer model and compare the performance between the complete testset and its hallucination part. The experimental results on various NMT datasets show that deficient, especially encoder embedding, and vulnerable cross-attentions are shown to be highly connected with hallucinations, while interestingly, cross-attentions provide good inductive bias and mitigate errors caused by encoder layers. 

\bibliography{neurips_2021}
\bibliographystyle{acl_natbib}

\clearpage
\appendix

\section{Multi-Domain and English-French Datasets}
To verify multi-domain scenarios, we use a multi-domain dataset \cite{koehn2017six,aharoni2020unsupervised} that contains five diverse domains (Medical, Law, Koran, IT, Subtitles) and each domain has 2000 valid and test sentence pairs.

To verify our findings on different languages, we use WMT'14 English-French as our in-domain dataset and IWSLT'14 English-French as our out-of-domain dataset. The WMT'14 English-French comprises 35.7M preprocessed sentence pairs, which are tokenized using the moses tool \footnote{\url{https://github.com/moses-smt/mosesdecoder/blob/master/scripts/tokenizer/tokenizer.perl}} and split using BPE. We use a combination of \emph{newstest2012} and \emph{newstest2013} as our validation set and \emph{newstest2014} as our test set, which consist of 6003 and 3003 sentences. The IWSLT'14 English-French test set is preprocessed using the same tokenization and BPE codes as the WMT'14 dataset. The testset combines IWSLT test sets from 2010 to 2014, with 6824 sentences. 

\section{Statistics of Hallucinated Samples}
\label{sec:stat}
In this section, we provide the statistics of all testsets, including their hallucinated parts, for a clearer understanding of hallucination results. The statistics are listed in Table \ref{table:stat}.

As we can see in our table, the number of hallucinated samples differs across various datasets and, especially, various domains. For instance, let us take a look at the "MD En-De",  LAW suffers much less hallucinations compared with other four domains. This diversity of datasets supports the generalization of our experiments. 

\section{Multi-domain Experiments}
\label{appendix:md5}
In the above experiments, we show several interesting findings in IWSLT dataset. However, there are concerns whether these findings are just working in one single dataset or domain. In this section, we conduct experiments over five diverse domains to support previous findings.

As shown in Table \ref{table:md5}, we can see that even though the decreases in encoder's word translation abilities vary in numbers, they demonstrate similar trends across five domains. 
The average accuracy difference from "All" to "Hallu." are -15.2 for embedding layer and -16.5 for 6-th encoder layer.
This verify our important finding in previous experiments that the main cause of hallucination is the weak translation ability of the encoder embedding in hallucinated inputs. 

\begin{table}[t]
\centering
\setlength\tabcolsep{5pt}
\begin{tabular}{c|c|r |r }
\toprule
\multicolumn{2}{c|}{Dataset} & \multicolumn{1}{c|}{Valid} & \multicolumn{1}{c|}{Test} \\
\midrule
\multicolumn{2}{c|}{IWSLT En-De} &746/8417 &652/7883 \\
\midrule
\multirow{5}{*}{MD En-De} &IT&865/2000& 868/2000 \\
&KORAN&516/2000 &510/2000 \\
&LAW&92/2000&121/2000 \\
&MEDICAL&385/2000&355/2000 \\
&SUBTITLES&662/2000&679/2000 \\
\midrule
\multicolumn{2}{c|}{IWSLT En-Fr} &460/9489 &416/6824 \\
\bottomrule
\end{tabular}
\caption{Statistics of all out-of-domain datasets we used. "X/Y" represents that X sentence pairs are identified as hallucinated samples in a dataset with a total of Y samples. "MD" stands for Multi-Domain Datasets.}
\label{table:stat}
\end{table}

\begin{table*}[h]
\centering
\setlength\tabcolsep{5pt}
\begin{tabular}{c||c |c |c |c |c |c |c|c|c|c|c}
\toprule
\multirow{2}{*}{\bf{Layer}} & \multicolumn{2}{c|}{\bf IT}  & \multicolumn{2}{c|}{\bf KORAN}  & \multicolumn{2}{c|}{\bf LAW} & \multicolumn{2}{c|}{\bf MEDICAL} & \multicolumn{2}{c|}{\bf SUBTITLES} & \multirow{2}{*}{\bf{$\Delta$}}\\
\cmidrule{2-11}
& {All} & {Hallu.} & {All} & {Hallu.} & {All} & {Hallu.} & {All} & {Hallu.} & {All} & {Hallu.} \\
\midrule
Emb. & 32.7 & 17.4 & 18.3 & 13.7 & 36.2 & 14.8 & 40.1 & 17.2 & 33.3 & 21.4 & -15.2 \\
\midrule
1& 34.7 & 23.3 & 18.3 & 13.7 & 35.3 & 17.1 & 39.6 & 18.2 & 34.0 & 21.8 & -13.6 \\
2& 35.6 & 23.1 & 18.9 & 14.1 & 36.5 & 17.0 & 40.7 & 18.3 & 34.3 & 22.1 & -14.3 \\
3 & 36.1 & 23.5 & 19.7 & 14.5 & 37.6 & 17.2 & 41.8 & 18.7 & 35.6 & 22.7 & -14.8 \\
4 & 36.9 & 23.8 & 20.5 & 15.0 & 38.7 & 17.5 & 42.9 & 19.3 & 36.5 & 22.7 & -15.4 \\
5 &37.7 & 23.7 & 21.0 & 15.3 & 40.1 & 17.8 & 43.7 & 19.6 & 37.0 & 22.9 & -16.0 \\
6 &37.5 & 22.9 & 21.5 & 15.1 & 40.6 & 18.0 & 43.7 & 19.0 & 37.6 & 23.3 & -16.5 \\
\bottomrule
\end{tabular}
\caption{Multi-domain experiments for word translation over hallucinations. "\textbf{Hallu.}" stands for the hallucination parts of each dataset, and "\textbf{All}" denotes the complete testsets. All results are accuracies and range from 0\% to 100\%. $\Delta$ denotes the average difference between "All" and "Hallu." among five different domains.}
\label{table:md5}
\end{table*}

\section{English-French Experiments}
\label{appendix:en-fr}
In this section, we provide our English to French experiments. We report 1-gram BLEU scores and accuracies, as discussed in Sec. \ref{sec:iwslt} and \ref{sec:w/o_cross}. Note that "1-BLEU" is the score with only encoder involved, and "Acc." is the translation score with both encoder and cross-attentions. 

As shown in Table \ref{table:enfr}, our findings are further confirmed in English-French. Firstly, both $\Delta$s for 1-BLEU scores and accuracies in En-Fr confirm the importantness of embedding layer in hallucination. Secondly, taking a look at "6-Emb.", 1-BLEU gaps enlarge with the increase of encoder layers, but accuracies do not change much. This validates our findings in the previous sections that, to some extent, cross-attentions mitigate hallucinated errors caused by encoder layers.
In conclusion, the experiments in English-French are consistent with our En-De results, proving the generalization of our findings. 

\begin{table*}[h]
\centering
\setlength\tabcolsep{5pt}
\begin{tabular}{c||c | c| c | c | c | c | c | c }
\toprule
\multirow{2}{*}{\bf{Layer}} & \multicolumn{2}{c|}{\bf WMT'14}  & \multicolumn{2}{c|}{\bf IWSLT All}  & \multicolumn{2}{c|}{\bf IWSLT Hallu.} & \multirow{2}{*}{\bf{$\Delta$ in 1-BLEU}} & \multirow{2}{*}{\bf{$\Delta$ in Acc.}}\\
\cmidrule{2-7}
& 1-BLEU & Acc. & 1-BLEU & Acc.& 1-BLEU & Acc.&  & \\
\midrule
Emb. & 23.3 & 41.7 & 25.1 & 40.6 & 8.2 & 26.8 & 16.9 & 13.8 \\
\midrule
1 &33.5 & 42.1 & 35.0 & 40.9 & 13.3 & 27.9 & 21.7 & 13.0 \\
2 &34.0 & 43.2 & 36.0 & 41.7 & 12.4 & 29.3 & 23.6 & 12.4 \\
3 &36.2 & 45.1 & 38.7 & 44.0 & 14.9 & 30.1 & 23.8 & 13.9 \\
4 &37.3 & 46.3 & 39.6 & 45.2 & 14.3 & 30.8 & 25.3 & 14.4 \\
5 &38.1 & 47.5 & 39.9 & 46.1 & 14.0 & 30.8 & 25.9 & 15.3 \\
6 &38.4 & 48.6 & 40.0 & 46.8 & 12.9 & 30.9 & 27.1 & 15.9 \\
\midrule
6 - Emb. & 15.1 & 6.9 & 15.0 & 6.2 & 4.7 & 4.1 & 10.3 & 2.1 \\
\bottomrule
\end{tabular}
\caption{English to French experiments for word translation over WMT'14 En-Fr, IWSLT'14 En-Fr All and IWSLT'14 En-Fr Hallucination datasets. "1-BLEU" represents the 1-gram BLEU score and is computed without cross-attention. "Acc." represents the word translation accuracy and is computed with cross-attention.}
\label{table:enfr}
\end{table*}



\end{document}